\pdfoutput=1

\documentclass[11pt]{article}

\usepackage{EMNLP2023}

\usepackage{times}
\usepackage{latexsym}

\usepackage[T1]{fontenc}

\usepackage[utf8]{inputenc}

\usepackage{microtype}

\usepackage{inconsolata}

\usepackage{amsmath,amsfonts,bm}

\def\eqref#1{equation~\ref{#1}}

\def\1{\bm{1}}

\def\mH{{\bm{H}}}

\def\mP{{\bm{P}}}

\def\wE{{\text{E}}}
\def\wF{{\text{F}}}

\def\wX{{\text{X}}}
\def\wY{{\text{Y}}}

\def\sentG{{G}}

\def\sentX{{X}}
\def\sentY{{Y}}

\DeclareMathAlphabet{\mathsfit}{\encodingdefault}{\sfdefault}{m}{sl}
\SetMathAlphabet{\mathsfit}{bold}{\encodingdefault}{\sfdefault}{bx}{n}

\newcommand{\R}{\mathbb{R}}

\newcommand{\prob}{\mathcal{P}}

\newcommand{\anno}[1]{\textcolor{blue}{#1}}
\newcommand{\enc}{\ensuremath{\mathtt{Encoder}}\xspace}
\newcommand{\dec}{\ensuremath{\mathtt{Decoder}}\xspace}

\newcommand{\add}{\ensuremath{\mathtt{add}}\xspace}

\newcommand{\eos}{\ensuremath{\langle\text{eos}\rangle}\xspace}
\newcommand{\bos}{\ensuremath{\langle\text{bos}\rangle}\xspace}
\newcommand{\sep}{\ensuremath{\langle\text{sep}\rangle}\xspace}

\newcommand{\delay}{\ensuremath{\mathtt{\Delta t}}\xspace}

\newcommand{\submax}{\ensuremath{\text{sub}^\text{max}}\xspace}
\newcommand{\timemax}{\ensuremath{\text{time}^\text{max}}\xspace}

\usepackage{wrapfig}

\usepackage{arydshln}

\usepackage{algorithm}
\usepackage{algorithmicx}
\usepackage[noend]{algpseudocode}

\usepackage{bibentry}
\usepackage{graphicx}
\usepackage{subfig} 
\usepackage{subfloat}
\usepackage{adjustbox}
\usepackage{multirow}
\usepackage{color}
\usepackage{xcolor}
\usepackage{booktabs}
\usepackage{amsmath}
\usepackage{amssymb}
\usepackage{colortbl}

\usepackage{subfig}
\usepackage{xspace}
\usepackage{makecell}
\usepackage{dblfloatfix}
\usepackage{float}
\usepackage{enumitem}
\usepackage{longtable}
\usepackage{bm}
\usepackage[english]{babel}
\usepackage{ragged2e}
\usepackage{tablefootnote}
\newcommand{\tabincell}[2]{\begin{tabular}{@{}#1@{}}#2\end{tabular}}

\usepackage{booktabs}
\usepackage{tabu}

\title{Pipelined Decoder for Efficient Context-Aware Text Generation}

\author{
Zixian Huang$^{1}$ \quad
Chenxu Niu$^{1}$ \quad
Yu Gu$^{2}$ \quad
Gengyang Xiao$^{1}$ \quad
Xinwei Huang$^{1}$ \quad
Gong Cheng$^{1}$\thanks{* Corresponding author.} \\
$^{1}$State Key Laboratory for Novel Software Technology, Nanjing University, Nanjing, China \\
$^{2}$The Ohio State University, Columbus, USA \\
\texttt{\{zixianhuang, cxniu, gyxiao, xwhuang\}@smail.nju.edu.cn} \\
\texttt{gcheng@nju.edu.cn}, \quad \texttt{gu.826@osu.edu}
}

\begin{document}
\maketitle

\begin{abstract}

As the basis of generative AI, an autoregressive model requires the generation of a new token depending on all the previously generated tokens, which brings high quality but also restricts the model to generate tokens one by one, forming a bottleneck limiting the generation speed. In this paper, we propose a new decoder architecture that efficiently generates text in parallel for context-aware generation tasks. Our proposed pipelined decoder initiates the generation of multiple subsequences simultaneously, and, at each time-step, it generates a new token for each subsequence to realize parallelism. Experiments on multiple text generation tasks, including question answering, text summarization, and keyphrase generation, show that our pipelined decoder significantly improves the generation speed without a significant loss of generation quality or additional memory consumption. \footnote{Our code is available on \url{https://github.com/HuangZixian/PipelinedDecoder}}

\end{abstract}
\section{Introduction}

The impressive text generation capabilities of generative models have ushered in a new era of artificial intelligence, positioning them as indispensable tools that greatly enhance productivity in various fields~\cite{GPT-4,Plan_Web_Agents}. As the adoption of these models continues to expand and user demand increases, optimizing their \emph{generation efficiency} has become a critical area of focus~\cite{speculative_decoding,Medusa,APAR}.
A major challenge is the quadratic time complexity of the self-attention mechanism, which scales with the length of the input text and creates a substantial efficiency bottleneck.

\begin{figure*}[t]
    \centering
    \includegraphics[width=1\textwidth]{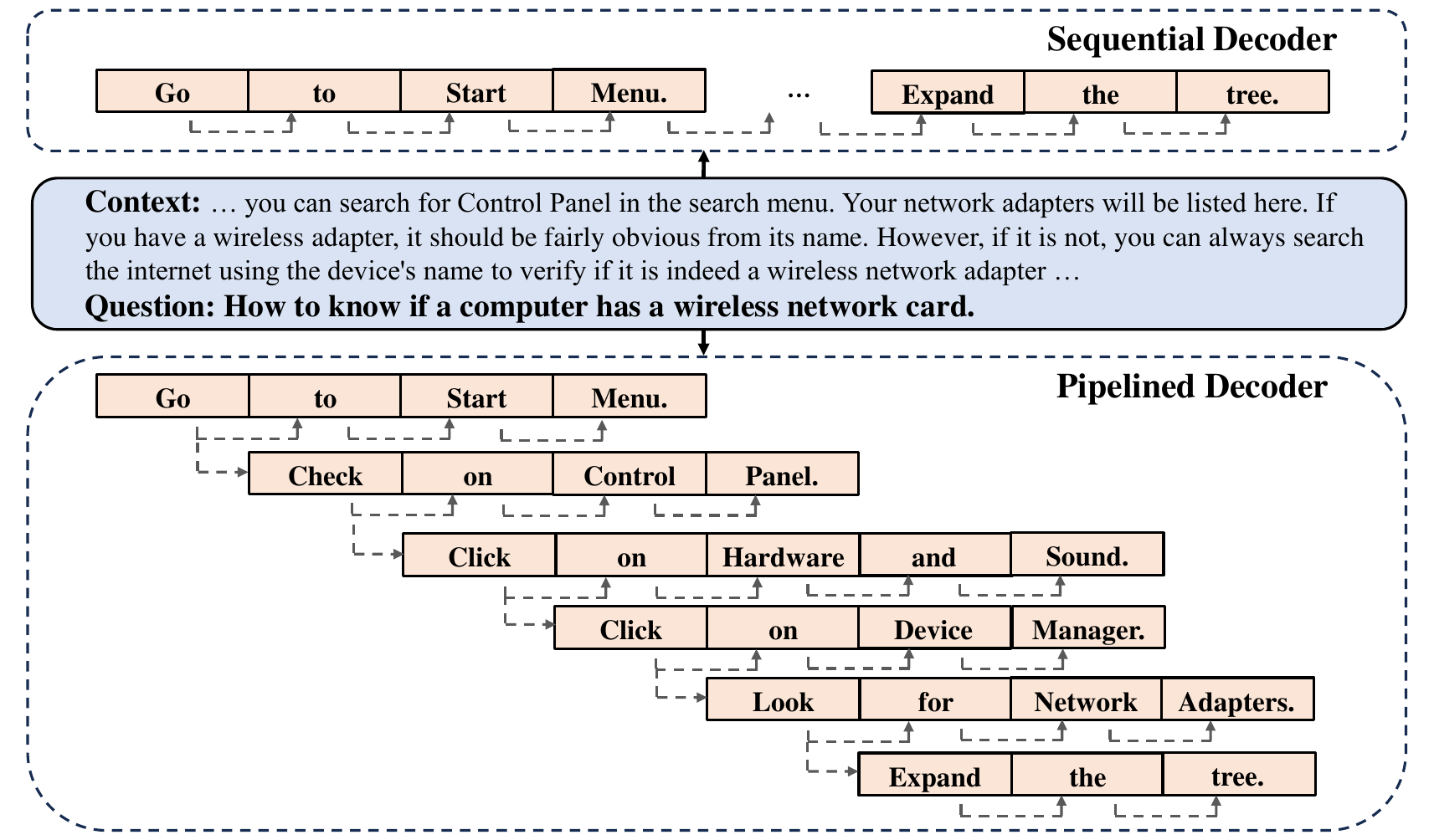}
    \caption{Comparison between the traditional sequential decoder and our pipelined decoder on an example of context-aware question answering from the WikiHowQA dataset.
    While the sequential decoder generates tokens serially, our pipelined decoder generates multiple subsequences in parallel.}
    \label{fig:case}
\end{figure*}

In particular, the challenge of efficiency becomes even more pronounced in \emph{context-aware} text generation~\cite{wikihowqa,PubMed,kptimes}, such as retrieval-augmented generation, text summarization, and keyphrase generation, where models must process extensive input contexts when generating new context.
An example is shown in Figure~\ref{fig:case}, where the model generates each answer token based on the given question and the retrieved relevant context. 
Additionally, existing generative models employ sequential decoders that generate tokens autoregressively, where each token depends on previously generated tokens, resulting in inherent computational latency.

\paragraph{Motivation.}
We argue that \emph{token generation does not inherently require strict autoregressive dependence on all previously generated tokens}. 
Intuitively, when writing a new sentence, humans rarely review all the words from previously written sentences but only need to keep an outline in mind.
Studies also show that generation quality is mainly affected by a few key tokens, such as the first token~\cite{stream-llm}.
Moreover, in context-aware generation, attention spans both input context and generated tokens, not just the latter.

\paragraph{Our Approach.}
Based on this observation, we consider \emph{whether the decoder can begin generating the next subsequence using only the initial few tokens from previous subsequences instead of the complete sequence}.
With this idea, we propose a \emph{pipelined decoder}, which starts the generation of multiple subsequences one after another in a pipelined manner. 
At a fixed delay time, the pipelined decoder starts generating a new subsequence. Multiple subsequences are generated in parallel, with each token attending to all context tokens and those generated in previous subsequences.


The lower part of Figure~\ref{fig:case} illustrates our pipelined decoder. Instead of depending on the complete previous subsequence ``Go to Start Menu'' as in the sequential decoder, our pipelined decoder generates the first token ``Check'' of the second subsequence only depending on the first token ``Go''.
This is feasible as recent studies have revealed that the hidden states of a token can encode information about current and future tokens~\cite{Medusa,meta_multi_token}, enabling the generation of multiple subsequences in parallel.
In this example, the pipelined decoder generates up to 4 tokens in parallel. As a result, while the sequential decoder requires 24 time-steps to generate the complete answer, the pipelined decoder only uses 8 time-steps.


\paragraph{Applications.}
Most text data can be viewed as ``long'' text composed of phrases or sentences as subsequences, so our pipelined decoder can be easily applied to most text generation tasks. 
We evaluate the performance of phrase- and sentence-level pipelined decoders on three context-aware text generation tasks: question answering, text summarization, and keyphrase generation.
Experimental results show that the pipelined decoder not only boasts a faster inference speed but also delivers generation quality comparable to that of the sequential decoder.


\section{Related Work}

\subsection{Generative Models}
Research on generative models has made great progress in recent years.
Through the exploration of different model architectures~\cite{T5,GLM}, the development of advanced generation strategies~\cite{Plan_And_Write,Progressive_Generation}, the enhancement of model understanding capabilities~\cite{mindmerger,dec2enc}, the mining of implicit knowledge within models~\cite{genmc,FBPrompt}, and scaling up model parameters~\cite{GPT-4}, the performance of generative models has improved substantially.
However, they use a sequential decoder that relies on the autoregressive (AR) model and generate tokens one by one, resulting in a bottleneck in the generation speed.

To obtain a faster generation speed, non-autoregressive~(NAR) model is proposed~\cite{NAR}, which ignores the dependence between sequences and generates tokens in parallel. 
Although the NAR model has achieved better performance based on better model architectures~\cite{semi-nar,Markov-transformers}, objective functions~\cite{NAR-Alignments,nar-edit-loss}, and learning paradigms~\cite{PLM-NAR,NAR-Curriculum}, independent token generation leads to generation quality often behind AR models~\cite{BANG}. In contrast, our pipelined decoder is capable of not only generating subsequences in parallel to \emph{accelerate inference}, but also partially depending on the previous subsequences to generate the subsequent subsequences to \emph{achieve generation quality comparable to that of the sequential decoder}.

\subsection{Inference Acceleration}
There are several categories of work that accelerate the inference of the AR model.
Some work optimizes operations such as I/O access from a system perspective~\cite{FlashAttention,Deepspeed}. Some work reduces the memory bandwidth cost by pruning key and value heads in the attention module~\cite{Multi-query-attention,kv_cache_h2o}. Quantization is another widely used solution to reduce memory consumption~\cite{GPTQ,SmoothQuant}.
Similar to our pipelined decoder, Reformer~\cite{Reformer} relies on some instead of all the previous tokens to generate new tokens to reduce computational cost. 
Unlike the above strategies, which still generate tokens serially one by one, our pipelined decoder achieves acceleration by \emph{generating multiple tokens in parallel}.

Some other work accelerates AR model inference through parallel generation.
Speculative decoding~\cite{speculative_decoding,speculative_sampling,Medusa} first uses a small model to generate a sequence of tokens, and then uses a large target model to refine the generated tokens in parallel, which is orthogonal to our pipelined decoder. 
Skeleton-based methods~\cite{SOT,APAR} first generate some skeleton points and then complete each skeleton point in parallel. However, not all target text can be clearly divided into structured skeleton points, while our pipelined decoder can, for example, \emph{simply rely on punctuation marks to divide the target text into multiple sentences} and generate them in parallel.
The most similar work to ours is the branching decoder~\cite{branching} proposed recently, which generates an unfixed number of subsequences in parallel, but is limited to set generation tasks. In contrast, our pipelined decoder is \emph{suitable for more diverse text generation tasks}.

\section{Pipelined Decoder}

\begin{figure}[t]
    \centering
    \includegraphics[width=1\linewidth]{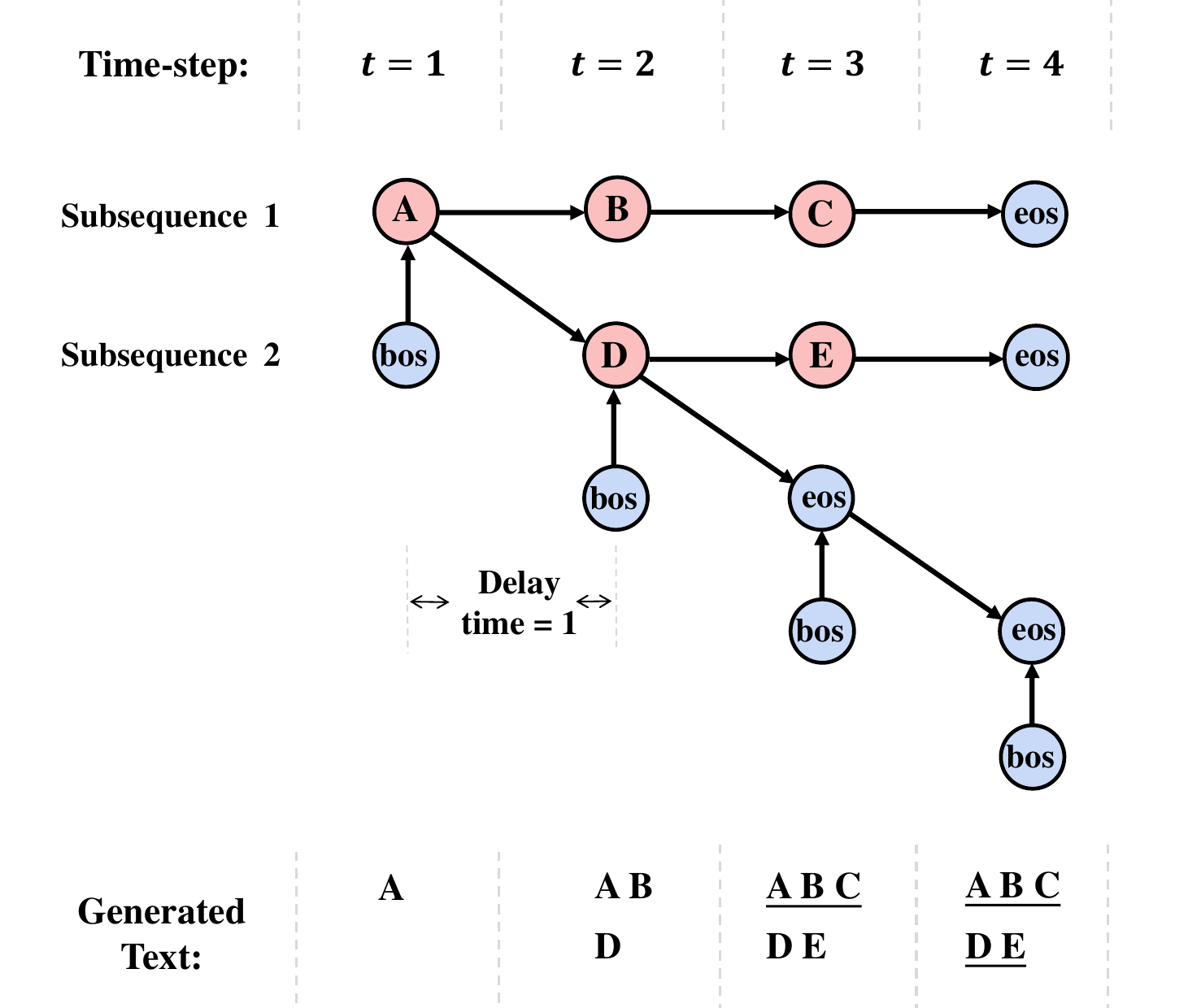}
    \caption{An example of the pipelined decoder for generating a sequence ``A B C \sep D E'', where \sep is a separator. 
    The pipelined decoder generates two subsequences ``A B C'' and ``D E'' in parallel, and only costs 4 time-steps to generate the complete sequence under the setting of \text{delay time} = 1.}
    \label{fig:decoding}
\end{figure}

\subsection{Overview}

Given an input sequence $\sentX={\wX_1, \ldots, \wX_l}$ with $l$ tokens, the goal of text generation is to generate a target sequence $\sentY=\sentY_1, \ldots, \sentY_n$ containing $n$ subsequences, where each subsequence $\sentY_i = \wY_{i,1}, \dots, \wY_{i, m_i}$ contains $m_i$ tokens.
Unlike a traditional sequential decoder that generates tokens one by one from left to right, the pipelined decoder \emph{generates multiple subsequences in parallel}. 

Figure~\ref{fig:decoding} illustrates pipelined decoding. To generate the sequence ``A B C $\sep$ D E'', where two subsequences ``A B C'' and ``D E'' are separated by a separating token $\sep$, the pipelined decoder simultaneously generates them with one time-step delay. Unlike a sequential decoder that generates 1~token at each time-step, the pipelined decoder generates 1, 2, 2 tokens, respectively, at time-steps 1 to 3.

Specifically, at time-step 1, the pipelined decoder generates the token ``A'' in subsequence 1.
After one delay in the time-step, at time-step 2, it starts the generation of subsequence 2 and generates ``B'' and ``D''. At time-step 3, it simultaneously generates ``C'' and ``E'' on subsequence 1 and subsequence 2, respectively, and explores the establishment of a new subsequence but fails due to the end token $\eos$ being generated. Finally, at time-step 4, subsequence 1, subsequence 2, and new subsequence exploration all generate $\eos$, which denotes the end of pipelined decoding.

\begin{figure*}[htbp]

\centering

\subfloat[Sequential Decoder] 
{ 
\includegraphics[width=0.4\linewidth, height=4.5cm]{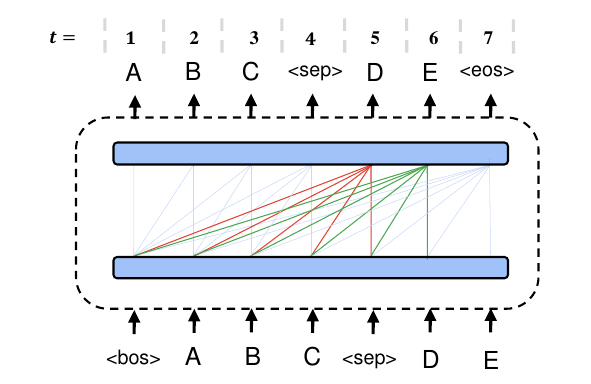}
}
\hfill 
\subfloat[Pipelined Decoder] 
{ 
\label{fig:pipe_training}
\includegraphics[width=0.5\linewidth, height=4.5cm]
{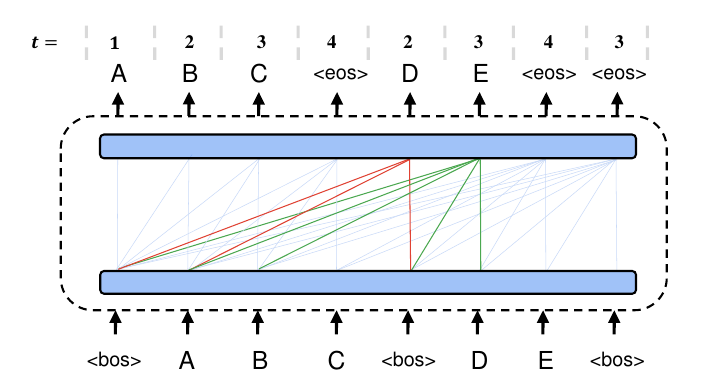}
}
\caption{Comparison between the sequential decoder and the pipelined decoder in attention distribution for modeling the sequence ``A B C \sep D E".
The attention distributions of ``D'' and ``E'' are highlighted in red and green, respectively.
}
\label{fig:training}
\end{figure*}

\subsection{Architecture}
Like a sequential decoder, the pipelined decoder uses a stacked transformer encoder to encode the input sequence $\sentX$:
\begin{equation}
\label{eq:enc}
      \mH^\wE = \enc(\sentX) \,.
\end{equation}
\noindent The hidden states of $\mH^\wE \in \R^{l \times d}$ will be fed into the pipelined decoder, where $d$ denotes the dimension of representation.

As a hyperparameter, the \emph{delay time} $\delay$ is set to control the pipelined decoder to establish a new subsequence every $\delay$~time-steps.
At time-step $t$, the current sequence $\sentG$ generated by the pipelined decoder is as follows:

\begin{align}
\label{eq:dec_input}
\begin{split}
\small
& \sentG = \sentG_{1},  \dots, \sentG_{k} \\
& \text{where} \quad k = \lceil t / \delay \rceil \,,
\end{split}
\end{align}
\noindent i.e.,~$\sentG$ consists of $k$ subsequences.
The $j$-th token in the $i$-th subsequence is generated at time-step $(i-1) \times \delay + j$.

Given the generated sequence $\sentG$ at time-step $t$ and the input hidden states $\mH^\wE$, the pipelined decoder generates $k$ new tokens for the $k$ subsequences in parallel at time-step $t+1$:
\begin{align}
\label{eq:dec}
\small
    \begin{split}
    \sentG_1^{t+1},\dots,\sentG_k^{t+1} &= \dec(\sentG, \mH^\wE) \,,
    \end{split}
\end{align}
where $\dec(\cdot, \cdot)$ is a stacked transformer decoder, which has the same parameters and architecture as a sequential decoder, but \emph{differs in the dependency between tokens}. Specifically, unlike a sequential decoder where
the generation of a new token at time-step $t+1$
depends on previous $t$ continuous tokens, a token in the $i$-th subsequence generated by the pipelined decoder depends on the first $i$ subsequences generated until time-step $t$, denoted by $G^{\leq t}_{\leq i}$, where \emph{the subsequences may be incomplete}. 
As shown in Figure~\ref{fig:decoding}, for the token ``D'', unlike a sequential decoder where ``D'' depends on five previous tokens ``\bos A B C \sep'', in the pipelined decoder it depends only on ``A'' and two \bos tokens.

For the $i$-th subsequence, the generation probability distribution of a new token $\sentG_i^{t+1}$ at time-step $t+1$ is:
\begin{align}
\label{eq:probaility}
\small
    \begin{split}
      \mP_i^{t+1} = \prob(G^{t+1}_{i}|G^{\leq t}_{\leq i};\mH^\wE)
      \,,
    \end{split}
\end{align}
where $\mP_i^{t+1} \in \R^v$ and $v$ is the size of the vocabulary. The token with the highest probability is selected as $\sentG_i^{t+1}$.


\subsection{Training}
The pipelined decoder is trained to obtain the ability to establish and terminate the generation of a subsequence.
Specifically, we add the beginning token $\bos$ before the $i$-th target subsequence $\sentY_i$ and add the end token $\eos$ after $\sentY_i$ to construct the training target subsequence $\hat{\sentY}_i = [\bos, \sentY_i, \eos]$.

When a sufficient number of subsequences have been generated, to help the pipelined decoder learn to stop generating new subsequences, we add an empty subsequence $\sentY^\wF = [\bos, \eos]$ at the end of the sequence to indicate that no more subsequences should be generated. The complete training target sequence is:
\begin{align}
\label{eq:teacher_forcing}
\begin{split}
\small
 \hat{\sentY} = [\hat{\sentY}_1, \dots, \hat{\sentY}_n, \sentY^\wF] \,,
\end{split}
\end{align}
where $\hat{\sentY}$ contains $n+1$ subsequences.

With $\hat{\sentY}$, the following loss is to be minimized:
\begin{align}
\label{eq:loss}
\begin{split}
\small
 \mathcal{L} = - \sum_{(i,t)}  \log
 \prob(\hat{Y}^{t+1}_{i}|\hat{Y}^{\leq t}_{\leq i}; \mH^\wE) .
\end{split}
\end{align}

Figure~\ref{fig:pipe_training} illustrates our teacher forcing training, where the training target contains three subsequences $\hat{\sentY}_1 = [\bos, \text{A}, \text{B}, \text{C}, \eos]$, $\hat{\sentY}_2 = [\bos, \text{D}, \text{E}, \eos]$, and $\sentY^\wF = [\bos, \eos]$. 


\subsection{Decoding}

\begin{algorithm}[!h]
\caption{Decoding}
\label{alg:decoding}
\textbf{Input}: $\sentX$, $\delay$, $\timemax$, $\submax$
\begin{algorithmic}[1]
\State $\sentG \leftarrow []$ \anno{// list of generated subsequences}
\State $k \leftarrow 0$ \anno{// number of generated subsequences}
\State $\mH^\wE \leftarrow \enc(\sentX)$

\For{$t \leftarrow$ 1 to $\timemax$} \anno{// at most $\timemax$ time-steps}

\If{$(t-1) \bmod \delay == 0$}
\If{$k < \submax$} \anno{// at most $\submax$ subsequences}
\State $\sentG.\add([ \bos ])$, $k \leftarrow k + 1$
\EndIf
\EndIf

\State $\sentG_1^{t+1},\dots,\sentG_k^{t+1} \leftarrow \dec(\sentG, \mH^\wE)$

\For{$i \leftarrow$ 1 to $k$}
\If{$\sentG_i.\text{endWith}(\eos)$ is \text{False}} 
\State $\sentG_i.\add(\sentG_i^{t+1})$ \anno{// incomplete}
\EndIf
\EndFor

\If{$\sentG$.\text{count}$(\eos) == k$ } 

\State \textbf{Break } \anno{// all complete}
\EndIf

\EndFor

\State \textbf{return} $\sentG.\text{remove}(\bos, \eos)$ 
\end{algorithmic}
\end{algorithm}

We show the decoding process of the pipelined decoder in Algorithm~\ref{alg:decoding}. The pipelined decoder is executed iteratively to generate the complete sequence. After every $\delay$ time-steps, a new subsequence is established. Each subsequence starts with $\bos$ and ends with $\eos$. We use the hyperparameter $\timemax$ to bound the number of iterations and use $\submax$ to bound the number of generated subsequences to avoid excessive generation in extreme cases.

\begin{table}[t!]
\centering

\caption{Dataset statistics. 
QA, KP, and TS refer to question answering, keyphrase generation, and text summarization, respectively.
$|\text{Tgt}|$, $\#~\text{Subseq}$, and $|\text{Subseq}|$ refer to the average number of words per target, 
the average number of subsequences per target,
and the average number of words per target subsequence, respectively. }
\resizebox{1\linewidth}{!}{
\begin{tabular} {l c rr rrr}
    \toprule
    Dataset & Task & \#~Train  & \#~Test 
     & $|\text{Tgt}|$ & \#~Subseq & $|\text{Subseq}|$\\ 
    \midrule
    
   \rowcolor{lightgray!30} Phrase-level &&&&&& \\
    \quad MSQA & QA & 4.6~K & 653 & 8.7 & 2.9 & 3.0 \\
    \quad KP20K & KP & 509~K & 20~K & 10.6 & 5.3 & 2.0 \\
    \quad KPTimes & KP & 259~K & 20~K & 11.0& 5.0 & 2.2 \\

    \midrule
     
     \rowcolor{lightgray!30} Sentence-level &&&&&&\\
    \quad WikiHowQA & QA & 129~K  & 40~K & 74.5 & 7.8 & 9.6 \\
    \quad CNN/DMail & TS & 287~K  & 12~K &  54.6 & 3.8 & 14.4 \\
    \quad PubMed & TS & 120~K & 7~K & 207.7  & 6.9 & 30.0 \\
    
    \bottomrule
\end{tabular}
}
\label{table:statistics}
\end{table}


Specifically, for an input sequence $X$, we first use Equation~(\ref{eq:enc}) to obtain its representation $\mH^\wE$~(line 3), and then iterate up to $\timemax$ time-steps to generate subsequences in parallel~(lines 4--13). 
After every $\delay$ time-steps, a new subsequence starting with $\bos$ is established~(lines 5--7). The pipelined decoder generates $k$ new tokens for the $k$ subsequences in parallel~(line 8). 
We add the generated tokens to the corresponding incomplete subsequences~(lines 9--11).
If all the subsequences generate the end token $\eos$, the pipelined decoder will complete generation~(lines 12--13).
Finally, removing the $\bos$ and $\eos$ from all the subsequences, we obtain the final sequence~(line 14).

\section{Experimental Setup}
\label{sec:setting}

\begin{table*}

\centering
\small
\caption{Comparison between the sequential decoder and the pipelined decoder on the MSQA dataset.
}
\label{table:msqa}

\begin{tabular}{l ccccclc}
\toprule
& \multirow{2}{*}{Backbone} & \multicolumn{2}{c}{Dev} & \multicolumn{2}{c}{Test} & \multirow{2}{*}{\tabincell{c}{Throughput \\ (examples per second)}} & \multirow{2}{*}{\tabincell{c}{GPU \\ Memory}}\\
 \cmidrule{3-6}
 & \multirow{2}{*}{} & EM & PM & EM & PM & \multirow{2}{*}{} & \multirow{2}{*}{} \\

\midrule
Sequential & \multirow{2}{*}{T5-Base} & 72.2 & 84.8 & 72.3 & 84.9 & 5.4 & 1,802~M\\ 
Pipelined & \multirow{2}{*}{} & 72.5 & 85.2 & 71.5 & 84.3 & 9.3~(1.7x) & 1,798~M \\

\midrule

Sequential & \multirow{2}{*}{T5-Large} & 75.5 & 87.3 & 74.7 & 87.3 & 3.1 & 3,713~M\\
Pipelined & \multirow{2}{*}{} & 75.0 & 87.2 & 74.0 & 87.1 & 5.3~(1.7x) & 3,710~M \\
\bottomrule

\end{tabular}

\end{table*}

\begin{table*}[t]

\centering
\small
\caption{Comparison between the sequential decoder and the pipelined decoder on two keyphrase generation datasets.
}
\label{table:kp_results}

\begin{tabular}{l ccccclc}
\toprule
 & \multirow{2}{*}{Backbone} & \multicolumn{2}{c}{Present} & \multicolumn{2}{c}{Absent} & \multirow{2}{*}{\tabincell{c}{Throughput \\ (examples per second)}} & \multirow{2}{*}{\tabincell{c}{GPU \\ Memory}}\\
 \cmidrule{3-6}
 & \multirow{2}{*}{} & F1@5 & F1@M & F1@5 & F1@M & \multirow{2}{*}{} & \multirow{2}{*}{} \\

\midrule
\rowcolor{lightgray!30} \textbf{KP20k} &  & & & & & & \\
\quad Sequential & \multirow{2}{*}{T5-Base} & 33.7 & 39.2 & 2.2 & 4.2 & 2.5 & 1,714 M \\
\quad Pipelined & \multirow{2}{*}{} & 27.8 & 40.1 & 4.0 & 7.6 & 5.1~(2.0x) & 1,634~M \\

\midrule

\quad Sequential & \multirow{2}{*}{T5-Large} & 33.0 & 40.5 & 2.1 & 4.1 & 1.6 & 3,972~M \\
\quad Pipelined & \multirow{2}{*}{} & 28.6 & 41.4 & 3.9 & 7.5 & 3.2~(2.0x) & 3,813~M \\
\midrule

\rowcolor{lightgray!30} \textbf{KPTimes} & & & & & & & \\
\quad Sequential & \multirow{2}{*}{T5-Base} & 33.7 & 51.0 & 24.2 & 33.0 & 3.4 & 1,534~M  \\ 
\quad Pipelined & \multirow{2}{*}{} & 32.2 & 50.8 & 22.8 & 33.4 & 8.4~(2.5x) &  1,467~M\\

\midrule

\quad Sequential & \multirow{2}{*}{T5-Large} & 34.8 & 51.8 & 24.7 & 34.1 & 1.8 & 3,944 M \\
\quad Pipelined & \multirow{2}{*}{}& 32.6 & 51.6 & 22.8 & 34.0 & 4.4~(2.4x) & 3,814~M \\

\bottomrule

\end{tabular}

\end{table*}

\begin{table*}[t]

\centering
\small
\caption{Comparison between the sequential decoder and the pipelined decoder on three sentence-level datasets.
}
\label{table:abstract_results}
\resizebox{1\linewidth}{!}{

\begin{tabular}{l ccccccc lc}
\toprule
 & \multirow{2}{*}{Backbone} & \multicolumn{3}{c}{Dev} & \multicolumn{3}{c}{Test} & \multirow{2}{*}{\tabincell{c}{Throughput \\ (examples per second)}} & \multirow{2}{*}{\tabincell{c}{GPU \\ Memory}}\\
 \cmidrule{3-8}
 & \multirow{2}{*}{} & Rouge-1 & Rouge-2 & Rouge-L & Rouge-1 & Rouge-2 & Rouge-L &
 \multirow{2}{*}{} & \multirow{2}{*}{} \\

\midrule
\rowcolor{lightgray!30} \textbf{WikiHowQA} &&&&&&&&&  \\
\quad Sequential & \multirow{2}{*}{T5-Base}  & 39.5  & 16.7 & 38.3 & 35.9 &  14.1 & 34.9 & 0.8 & 3,782~M\\ 
\quad Pipelined & \multirow{2}{*}{} & 39.2 & 16.6 & 38.0 & 35.7 & 13.9 & 34.6 & 4.6~(5.8x) & 3,428~M\\

\midrule

\quad Sequential & \multirow{2}{*}{T5-Large} & 41.2 & 18.5 & 40.1 & 37.8 & 15.7 & 36.7 & 0.5 & 7,570~M \\
\quad Pipelined & \multirow{2}{*}{} & 40.9 & 18.4 & 39.7 & 37.3 & 15.6 & 36.2 & 2.6~(5.2x) & 5,994~M \\
\midrule

\rowcolor{lightgray!30} \textbf{CNN/DMail} &&&&&&&&& \\
\quad Sequential & \multirow{2}{*}{T5-Base} & 39.6 & 17.8 & 37.5 & 39.2 & 17.4 & 37.1 & 1.2 & 3,436~M\\
\quad Pipelined & \multirow{2}{*}{} & 38.4  & 17.2 & 36.4 & 37.8  & 16.6 & 35.8  & 2.9~(2.4x) & 3,334~M\\

\midrule

\quad Sequential & \multirow{2}{*}{T5-Large} & 39.6 & 17.8 & 37.5 & 39.2 & 17.4 & 37.1 & 0.7 & 6,170~M\\
\quad Pipelined & \multirow{2}{*}{} & 38.5 & 17.5 & 36.5 & 38.0 & 17.0  & 36.0 & 1.6~(2.3x) & 5,874~M\\

\midrule

\rowcolor{lightgray!30} \textbf{PubMed} & &&&&&&&& \\
\quad Sequential & \multirow{2}{*}{T5-Base} & 39.8 & 18.3 & 36.9 & 39.7 & 18.2 & 36.8 & 0.2 & 4,514~M \\
\quad Pipelined & \multirow{2}{*}{} & 38.2  & 17.5 & 35.5 & 38.2 & 17.4 & 35.4  &  1.3~(6.5x) & 4,390~M \\

\midrule

\quad Sequential & \multirow{2}{*}{T5-Large} & 40.5 & 18.9 & 37.6 & 40.5 & 18.7 & 37.6 & 0.1 & 6,908~M \\
\quad Pipelined & \multirow{2}{*}{} &  38.7  & 17.8 & 35.8  & 38.4 & 17.9 & 36.2 & 0.7~(7.0x) & 6,482~M\\




\bottomrule

\end{tabular}
}

\end{table*}

\subsection{Datasets}
We evaluated the pipelined decoder on \emph{phrase-level} and \emph{sentence-level context-aware text generation datasets} to verify its effectiveness and generalizability, as shown in Table~\ref{table:statistics}.

For phrase-level datasets, we selected one \emph{question answering} dataset \textbf{MSQA}~\cite{msqa} and two \emph{keyphrase generation} datasets \textbf{KP20K}~\cite{KP20K} and \textbf{KPTimes}~\cite{kptimes}.
MSQA is a multi-span question answering dataset that requires the model to generate multiple spans from a given document to answer a question~\cite{spanqualifier}.
KP20K and KPTimes are two large datasets from the science and news fields that require the model to generate both present and absent keyphrases from a given document.

For sentence-level datasets, we selected one question answering dataset \textbf{WikiHowQA}~\cite{wikihowqa} and two \emph{text summarization} datasets  \textbf{PubMed}~\cite{CNNDM} and \textbf{CNN/DMail}~\cite{PubMed}.
WikiHowQA is a community question answering dataset that requires the model to generate answers to relevant questions from a given document. CNN-DM and PubMed are two large datasets from the science and news fields that require the model to generate a summary of a given document.

\subsection{Evaluation Metrics}
For the MSQA dataset, we used Exact Match F1~(\textbf{EM}) and Partial Match F1~(\textbf{PM}) introduced by~\cite{msqa}, where EM requires an
exact match between the generated answer and the ground-truth answer, and PM generalizes EM by using the length of the
longest common substring.

For keyphrase generation, we used the macro-averaged \textbf{F1@5} and \textbf{F1@M} introduced by \cite{kp_metric} to report the generation quality of present and absent keyphrases. When the number of predicted keyphrases is less than 5, F1@5 first appends incorrect keyphrases to 5 and then compares with the ground truth to calculate an F1 score. F1@M is the version of F1@5 without appending incorrect keyphrases.

For all the sentence-level datasets, we used \textbf{Rouge} F1 to evaluate generation quality against the ground truth.

\subsection{Implementation Details}
We implemented the pipelined decoder and the sequential decoder based on the code of huggingface transformers 4.12.5 and used T5~\cite{T5} as the backbone. We set the hyperparameter delay time $\delay=1$ for maximum acceleration, and set $\timemax=50$ and $\submax=20$ to ensure that all subsequences can be generated completely. We used the Manhattan distance to calculate the relative position between two tokens for the attention layer of T5: for $\sentG^{t_1}_{i_1}$ and $\sentG^{t_2}_{i_2}$, their relative distance is $|t_1-t_2| + |i_1-i_2|$.

We used batch size $24$, learning rate $1e-4$, maximum sequence length $2048$, epoch number $10$, and the AdamW optimizer. We trained a pipelined decoder based on T5-Base~(223~M) on a single RTX 4090 (24~G), and trained a pipelined decoder based on T5-Large~(738~M) on two RTX 4090. For inference, we ran both base and large versions on a single RTX 4090 with batch size $1$.

\section{Experimental Results}
\label{sec:results}

\subsection{Results on Phrase-Level Datasets}
Table~\ref{table:msqa} and Table~\ref{table:kp_results} show the performance of the pipelined decoder on three phrase-level datasets, including its generation quality, inference speed (i.e., throughput) and GPU memory usage. Compared to the sequential decoder, the pipelined decoder not only has \emph{higher inference speed} and \emph{lower GPU memory usage}, but also, satisfyingly, demonstrates \emph{comparable generation quality}.

\paragraph{Inference Speed.} 
On the MSQA dataset, the pipelined decoder achieves a 1.7x speedup compared to the sequential decoder. It shows an even greater speedup on the KP20K and KPTimes datasets with T5-Base, achieving an improvement of 2.0x and 2.5x, respectively. This is because the average number of target subsequences in KP20K and KPTimes is larger than that in MSQA (see Table~\ref{table:statistics}), allowing the pipelined decoder to establish more parallel subsequences.

\paragraph{Generation Quality.}
Although the pipelined decoder only partially depends on the tokens in initial positions, its generation quality reflected by most metrics is not worse than that of the sequential decoder with full token dependencies.
In the question answering task on MSQA, compared to the sequential decoder, the largest decline of the pipelined decoder occurs in the EM score on the test set with T5-Base, with only a slight decrease of 0.8\%.
The pipelined decoder outperforms the sequential decoder in 8 of 16 metrics, particularly in the absent keyphrase prediction task on KP20K, with an improvement ranging from 1.8\% to 3.4\%.



\paragraph{GPU Memory Usage.} 
Compared to the sequential decoder, the pipelined decoder does not incur an additional computational cost and, since there are only partial token dependencies, the pipelined decoder uses even less memory. 

\subsection{Results on Sentence-Level Datasets}
Table~\ref{table:abstract_results} shows the performance of the pipelined decoder on three sentence-level datasets. It \emph{significantly improves inference speed} while \emph{using less GPU memory}, achieving \emph{generation quality comparable to that of the sequential decoder}.

\paragraph{Inference Speed.} 
On the WikiHowQA and PubMed datasets, the pipelined decoder achieves throughput at least 5.2x higher on WikiHowQA and 6.5x higher on PubMed than the sequential decoder.
In contrast, the average number of target subsequences in the CNN/DMail dataset is relatively small (see Table~\ref{table:statistics}), resulting in a relatively slow acceleration for the pipelined decoder, still improving by at least 2.3x.


\paragraph{Generation Quality.}
Unlike phrase-level short text generation, generating subsequences at the sentence level faces a higher risk of missing dependencies for the pipelined decoder.
In this regard, it is satisfactory that the pipelined decoder and the sequential decoder are close in most metrics, with the largest gap of only 0.5\%, 1.4\%, and 2.1\% on WikiHowQA, CNN/DMail, and PubMed, respectively.


\paragraph{GPU Memory Usage.}
Again, unsurprisingly, the pipelined decoder uses less memory than the sequential decoder.

\begin{figure}[t]
    \centering
    \includegraphics[width=\linewidth]{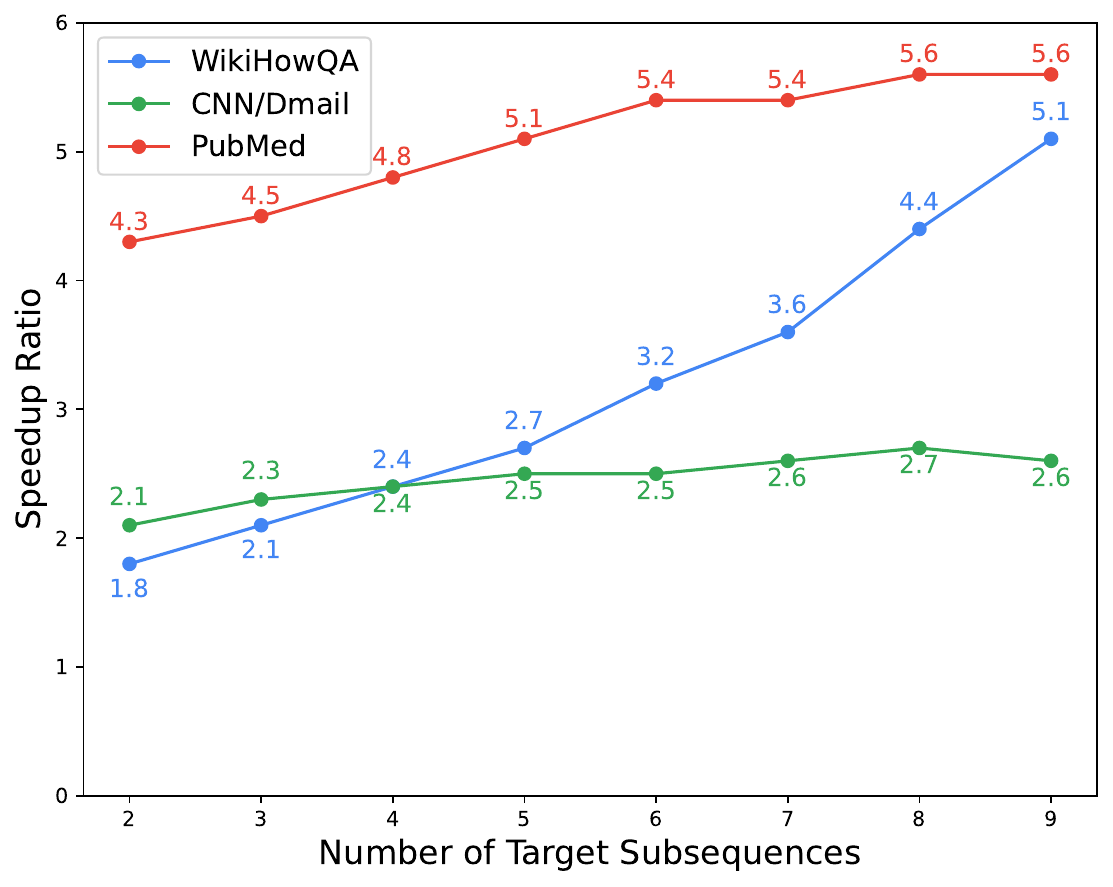}
    \caption{Speedup ratio of the pipelined decoder to the sequential decoder under different numbers of target subsequences.}
    \label{fig:speed_ratio}
\end{figure}

\subsection{Impact of the Number of Target Subsequences}

\paragraph{Impact on Inference Speed.}
The speedup of the pipelined decoder arises from its parallel generation of multiple subsequences. For longer text that can be divided into more subsequences, the pipelined decoder is expected to offer faster acceleration.
Figure~\ref{fig:speed_ratio} plots the relationship between the average number of target subsequences and the speedup ratio.
Indeed, \emph{when the number of target subsequences increases, the speedup ratio of the pipelined decoder to the sequential decoder increases}.
Specifically, as the number of target subsequences increases from 2 to 9, the speedup ratio increases from 1.8x to 5.1x on WikiHowQA, from 2.1x to 2.6x on CNN/DMail, and from 4.3x to 5.6x on PubMed.


\begin{figure}[t]
  \centering
 \includegraphics[width=\linewidth]{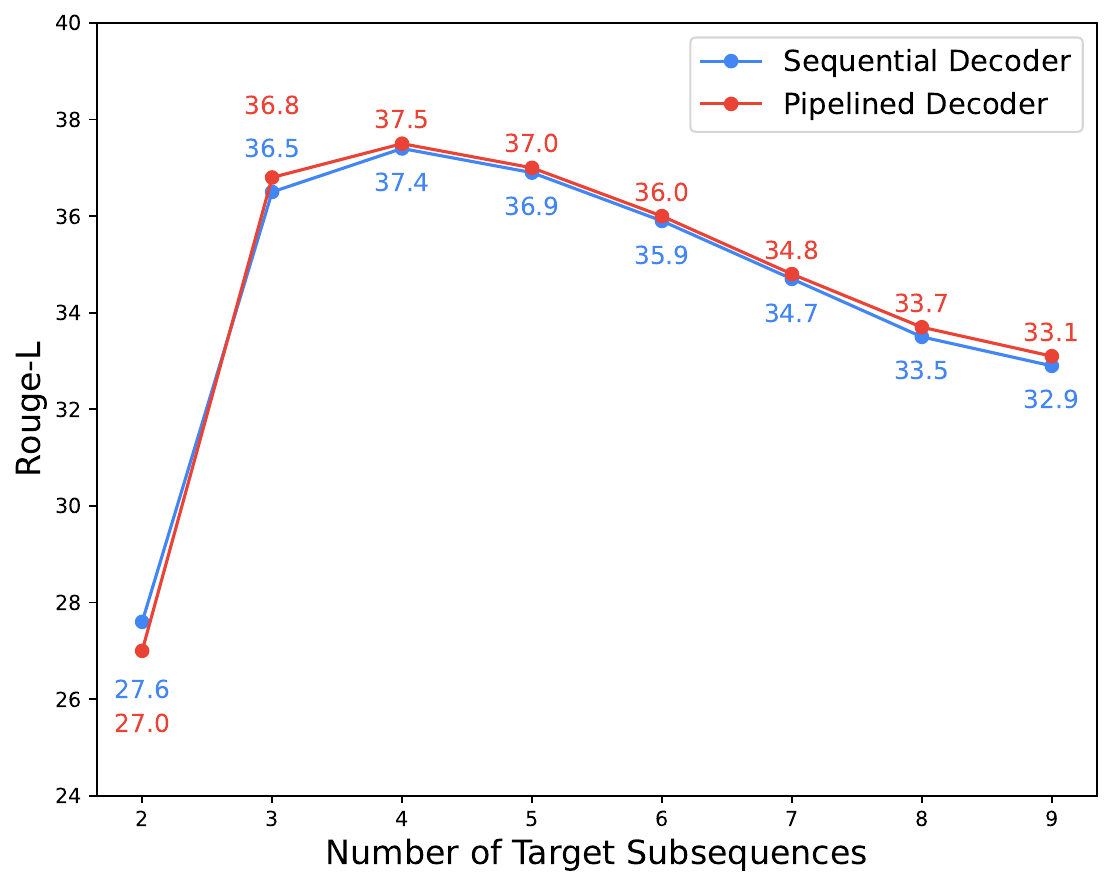}
  \caption{Comparison between the sequential decoder and the pipelined decoder in generation quality (i.e., Rouge-L) under different numbers of target subsequences on the WikiHowQA dataset.}
  \label{fig:subseq_num}
\end{figure}

\paragraph{Impact on Generation Quality.}
We further analyzed whether an increase in the number of target subsequences affects generation quality. We selected WikiHowQA for analysis, as according to Figure~\ref{fig:speed_ratio}, on this dataset the speedup ratio is the most sensitive to the change in the number of target subsequences. Figure~\ref{fig:subseq_num} plots the relationship between the average number of target subsequences and the generation quality. Except under 2~target subsequences, the Rouge-L score of the pipelined decoder consistently exceeds that of the sequential decoder by 0.1\% to 0.3\%, indicating that \emph{the increased speedup of the pipelined decoder does not come at the expense of generation quality}.

\subsection{Impact of Delay Time}
\begin{table}[t]

\centering
\small
\caption{Performance of the pipelined decoder under different values of delay time $\delay$ on the test sets of the three sentence-level datasets.}
\label{table:delay_time}
\resizebox{\linewidth}{!}{
\begin{tabular}{l ccc c}
\toprule
 & Rouge-1 & Rouge-2 & Rouge-L & \tabincell{c}{Throughput \\ (examples per second)}\\

\midrule
\rowcolor{lightgray!30} \textbf{WikiHowQA} & &&&\\

\quad $\delay=1$ & 35.7 & 13.9 & 34.6 & 4.6 \\ 
\quad $\delay=2$ & 35.9 & 13.9 & 34.8 & 4.4 \\ 
\quad $\delay=3$ & 35.8 & 13.7 & 34.7 & 4.3 \\ 

\midrule
\rowcolor{lightgray!30} \textbf{CNN/DMail} & &&& \\

\quad $\delay=1$ & 37.8 & 16.6 & 35.8 & 2.9\\ 
\quad $\delay=2$ & 37.5 & 16.5 & 35.5 & 2.7\\ 
\quad $\delay=3$ & 38.0 & 16.9 & 36.1 & 2.6\\ 

\midrule

\rowcolor{lightgray!30} \textbf{PubMed} &&&& \\

\quad $\delay=1$ & 38.2 & 17.4 & 35.4 & 1.3\\ 
\quad $\delay=2$ & 38.4 & 17.6 & 35.7 & 1.3 \\ 
\quad $\delay=3$ & 37.6 & 17.2 & 34.9 & 1.2 \\ 

\bottomrule

\end{tabular}
}

\end{table}

The pipelined decoder uses a hyperparameter delay time $\delay$ to control the frequency of generating new subsequences. Although a smaller $\delay$ starts the parallel generation earlier, it may cause subsequent subsequences to depend less on the previous subsequences. Therefore, the effect of this hyperparameter deserves investigation.

As shown in Table~\ref{table:delay_time}, when $\delay$ increases, the inference speed decreases as expected, while the generation quality fluctuates in a narrow range, suggesting that \emph{the information contained in the initial tokens of the subsequences is often sufficient to support the generation of the remaining text}.




\section{Conclusion}

In this paper, we propose a novel pipelined decoder that generates multiple subsequences in parallel, resembling a pipeline. In both phrase- and sentence-level context-aware text generation tasks, including question answering, text summarization, and keyword generation, the pipelined decoder significantly improves generation efficiency while preserving generation quality. Therefore, this novel generative architecture sees a wide range of promising applications.

\paragraph{Limitations.}
(1)~Despite the variety of context-aware tasks, the pipelined decoder may not be suitable for context-free tasks such as math word problems where it is essential to attend to previously generated tokens so that the partial token dependencies in the pipelined decoder are probably insufficient.
(2)~Due to resource limitations, we implemented the pipelined decoder only over T5 but not larger backbones. 

\paragraph{Future Work.}
The pipelined decoder presents many research problems that deserve in-depth exploration.
In the future, we will seek to establish new subsequences adaptively rather than at fixed time intervals. We will also extend the pipelined decoder with other existing acceleration strategies.






\end{document}